\documentclass{article}
\pdfoutput=1
\usepackage[english]{babel}
\usepackage[utf8]{inputenc}
\usepackage{csquotes}
\usepackage{amsmath,amsthm,amssymb,latexsym,epic,eepic,epsfig,graphics,psfrag}
\usepackage{amsfonts}
\usepackage[svgnames]{xcolor}
\usepackage{parskip}
\usepackage{bm}
\usepackage{subcaption}
\usepackage{float}
\usepackage{listings}
\usepackage{marginnote}
\usepackage{setspace}
\usepackage[hyphens]{url}             
\usepackage[unicode=false]{hyperref}                 
\usepackage[numbers,sort&compress]{natbib}
\usepackage{booktabs}

\newcommand{\figref}[1]{Fig.~\ref{#1}}
\newcommand{\secref}[1]{Section \ref{#1}}
\newcommand{\eqnref}[1]{Eq.~(\ref{#1})}

\theoremstyle{definition}

\theoremstyle{remark}

\numberwithin{equation}{section}

\newcommand\keywordstyle{\rmfamily\bfseries\upshape}

\lstdefinelanguage{pseudocode}{
  keywordstyle=[1]{\keywordstyle},
  keywords=[1]{%
    begin,end,%
    program,procedure,%
    while,do,for,to,break,continue,repeat,until,od,in,%
    if,then,else,fi,return,yield},
  morecomment=[l]{//},
  morecomment=[s]{/*}{*/}
}

\begin{document}

\title{\large{Technical report}\medskip\\\LARGE{Predicting clicks in online display advertising with latent features and side-information}}%

\author{Bjarne \O rum Fruergaard\\b.fruergaard@adform.com,\\Adform Aps, Hovedvagtsgade 6 th., 1103 K\o benhavn K, Denmark}%
\date{\today}%

\maketitle

\begin{abstract}
We review a method for click-through rate prediction based on the work of \citet{Menon2011}, which combines collaborative filtering and matrix factorization with a side-information model and fuses the outputs to proper probabilities in $[0,1]$. In addition we provide details, both for the modeling as well as the experimental part, that are not found elsewhere. We rigorously test the performance on several test data sets from consecutive days in a click-through rate prediction setup, in a manner which reflects a real-world pipeline. Our results confirm that performance can be increased using latent features, albeit the differences in the measures are small but significant.
\end{abstract}

\clearpage
\tableofcontents

\section{Introduction}
With the growing popularity of the Internet as a media, new technologies for targeting advertisements in the digital domain, a discipline generally referred to as \textit{computational advertising}, have opened up to new business models for publishers and advertisers to finance their services and sell their products. Online advertising entails using banner ads as a means to attract user attention towards a certain brand or product. The clicks, known as \textit{click-throughs}, take a user to a website specified by the \textit{advertiser} and generates revenue for the page displaying the banner, which we call the \textit{publisher}.

In \textit{real-time bidding} (RTB) banner ads are determined and placed in real-time based on an auction initiated by the publisher between all potential advertisers, asking them to place a bid of what they are willing to pay for the current \textit{impression} (displaying the ad), given information about the page, the user engaging the page, a description of the banner format and placement on the page. The advertiser with the highest bid wins the auction and their banner is displayed to the user. RTB thus requires advertisers, or more commonly, the \textit{demand side platform}s (DSPs) acting on behalf of the advertisers, to be able to estimate the potential value of an impression, given the available information. A key measure for evaluating the potential values of impressions is the \textit{click-through rate} (CTR), calculated as the ratio of the number of clicks over the total number of impressions in a specific context. What we are investigating in the present work, is a model for predicting CTRs, even in the face of contexts without any previous clicks and/or very few impressions available, such that the empirical CTR can be unknown or very poorly estimated.



\subsection{Dyadic prediction}
We frame our main objective of estimating click-through rates for web banner advertisements in the general scope of a \textit{dyadic prediction} task. Dyadic prediction concerns the task of predicting an outcome (or label) for a \textit{dyad}, $(i,j)$, whose members are uniquely identified by $i$ and $j$, but which may include additional attributes of the dyad $(i,j)$ being observed. 

In this paper we are interested in predicting the binary labels being either \textit{click} or \textit{not click}, in general referred to as \textit{click-through rate} prediction, given the pair of a \textit{domain} and a web \textit{banner} advertisement. In the following, we give a formal introduction of this problem.

We are given a transaction log of banner advertisements being shown to users. In the logs, various dimensions are recorded, including a \textit{banner ID} and a \textit{domain ID}, as well as a number of other attributes, which we shall elaborate more on later. For each record in the log, henceforth called a \textit{view}, it is recorded whether the banner was clicked or it was displayed without any subsequent click (non-click). Let $i=1,...,M$ index the banner dimension and $j=1,...,N$ the domain dimension. We can then construct a matrix, $\bm X$, summarizing the records in the log in terms of empirical click-through rates, i.e., let the entries of the matrix be defined by
\begin{align}
	X_{ij} &= \left\{ \begin{array}{ll}
		\frac{C_{ij}}{V_ij} & \text{if }V_{ij} > 0 \\
		? & \text{otherwise} \end{array} \right.
\end{align}
Here $C_{ij}$ is the number of clicks and $V_{ij}$ is the number of views involving dyad $(i,j)$. Note that per definition, both clicks and non-clicks count as views, so we must always have $V_{ij} \geq C_{ij}$. The ``?'' denotes unobserved pairs, where there is no historical data in the log, hence for such dyads $X_{ij}$ is undefined.

With this formulation, our click-through rate prediction task is to learn models estimating $\bm X$. Naturally, any such model should be able to predict the missing entries ``?'', as well as being able to \textit{smoothen} predictions, such that the model does not get over-confident in situations with too few views. For instance, if $C_{ij}=1$ and $V_{ij}=3$, a CTR estimate of $X_{ij}=\frac13$ is probably too extreme, as well as the case $C_{i'j'}=0 \Rightarrow X_{i'j'}=0$, where the natural assumption should rather be that not enough pairs $(i',j')$ have yet been observed.

One possible approach to the above is where additional features about the entities $i$ and $j$ are known. This \textit{side-information} can then be used as predictors in a supervised learning model, such as logistic regression. We refer to this approach as \textit{feature-based}.

In the complete lack of side-information, one can instead borrow ideas from \textit{collaborative filtering}. In collaborative filtering the classic setup, e.g., the Netflix movie rating problem \cite{Bennett2007}, is where dyads are (user,item) pairs and each observed pair is labeled with a rating, for instance on the scale 1 to 5. The task is then to predict the ratings for unobserved pairs, which can then be used as a component in a recommender system. In our case we can identify a similar collaborative filtering task, but where instead of ratings we have binary outcomes and the dyads are (banner,domain) pairs. The assumption in collaborative filtering is that for a particular (banner,domain) pair, other row objects (other banners) as well as other column objects (other domains) contain predictive information. I.e., we are assuming that some information is \textit{shared} between entities and we need to learn a model of this shared information.

In this work we investigate a model that fuses ideas from collaborative filtering via matrix factorization and a mapping to valid probabilities in $[0,1]$, called a \textit{latent feature log-linear model} (LFL) with a feature-based model for explicit features, that we refer to as a \textit{side-information} model.



\subsection{Related work}
The model that we investigate in this work was introduced in \cite{Menon2011} and builds on the latent feature log-linear model (LFL) from \cite{Menon2010}. Our work can be seen as a supplement to \cite{Menon2011}, as we think this work is lacking in details, which we thus try and provide. Also, we offer different conclusions about the applicability of this model to a dataset of our own, but for the same application as \cite{Menon2011}. \cite{Menon2011} does not share any of their data so we can unfortunately not reproduce their results.

The modeling of click-through rates has been extensively investigated in the domain of search engine advertising, i.e., the sponsored advertisements that appear in web search engines as a result of user queries for relevant content retrieval. Many methods proposed in this domain are feature-based, e.g., \cite{Richardson2007,Chakrabarti2008,Graepel2010} based on logistic regression. Other techniques are maximum likelihood based \cite{Dembczynski2008,Ashkan2009}, i.e., they operate directly with the empirically observed counts, which makes it a problem to predict in cold-start settings. Since in search engines, the user directly reveals an \textit{intent} through his or her query, the features in most of these studies include somehow to predict click-through rates of pairs of (word,ad), which could indeed also be modeled using the LFL framework \cite{Menon2010}, but to our knowledge this has yet to be investigated.

In the setting that we are looking at, namely placement of banner ads, there is no direct query from the user, so modeling click-through rates cannot be based on (word,ad) pairs and have to be based on other often much weaker predictors of intent. Feature-based approaches are also popular in this setting, see e.g. \cite{McMahan2013}. Latent feature models are also not much explored in this area, hence a motivation for this work is to combine the best of combining latent and explicit features and share our findings.

Our focus in this work is on combining the LFL model \cite{Menon2010} with a logistic regression model on the explicit features as in \cite{Menon2011}. This combined model has the advantage that it in the face of weak explicit predictors, recommender effects from the latent features can kick in.

\section{Response prediction}

The model we apply for response prediction is based on the work in \cite{Menon2011}, a collaborative filtering technique based on matrix factorization, which is a special case of the latent feature log-linear model (LFL) \cite{Menon2010} for dyadic prediction with binary labels. \citeauthor{Menon2011} demonstrate that their model incorporating side-information, hierarchies and an EM-inspired iterative refinement procedure overcome many collaborative filtering challenges, such as sparsity and cold-start problems, and they show superior performance to models based purely on side-information and most notably the LMMH model \cite{Agarwal2010}. In the following we introduce the \textit{confidence-weighted} latent factor model from \cite{Menon2011}.

\subsection{Confidence-weighted factorization}\label{sub:cwfact}
A binary classification problem of the probability of click given a dyadic observation $(i,j)$ for page $p_i$ and banner $b_j$, $p(click | (i,j))$, can be modeled with the \textit{logistic function} and a single weight, $\omega_{ij}$, per dyad. \textit{I.e.}, $p^{LR}(click | \omega_{ij}) = \sigma(\omega_{ij})$. However, such a model is only capable of classifying dyads already observed in training data and cannot be applied to unseen combinations of pages and banners. Therefore we assume a factorization of $\omega_{ij}$ into the factors $\bm \alpha_i$ and $\bm \beta_j$ each representing latent feature vectors of the page and banner dimensions, respectively, such that  $\omega_{ij} \approx \bm \alpha_i^T \bm \beta_j$. Henceforth, we will refer to this estimator as $p_{ij}^{MF} := p^{MF}(click | \bm \alpha_i, \bm \beta_j) = \sigma(\bm \alpha_i^T \bm \beta_j)$.

With data being $d=1,...,D$ observations of dyads, $x_d=(i,j)$, with binary labels, $y_d \in \{0,1\}$, learning can be formulated as the regular logistic regression optimization problem:
\begin{align}
    \min_{\bm A, \bm B} \quad - \sum_{d=1}^D y_d \log(p^{MF}_{i_dj_d}) + (1 - y_d) \log(1 - p^{MF}_{i_dj_d}), \label{eq:negloglike_ll}
\end{align}
\textit{i.e.}, a maximum-likelihood solution for Bernoulli-distributed output variables using the logistic function to non-linearly map continuous values to probabilities. With the latent variables $\bm A$ and $\bm B$ being indexed by $(i,j)$, we can rewrite \eqnref{eq:negloglike_ll} to a \textit{confidence-weighted factorization}:
\begin{align}
    \min_{\bm A, \bm B} \quad - \sum_{(i,j) \in \mathcal{O}} C_{ij} \log(p^{MF}_{ij}) + (V_{ij} - C_{ij})\log(1 - p^{MF}_{ij}), \label{eq:negloglike_cwf}
\end{align}
where $C_{ij}$ is the number of clicks ($y_d=1$) involving dyad $(i,j)$ and $(V_{ij}-C_{ij})$ the number of non-clicks ($y_d=0$) involving dyad $(i,j)$ in the training data. This reformulation can be a significant performance optimization, since the number of distinct dyads can be much smaller than the total number of observations. \textit{E.g.}, in the case of click-through data, we can easily have many thousands of click and (particularly) non-click observations per dyad, hence the number of operations involved in the summation of \eqnref{eq:negloglike_cwf} is significantly reduced compared to \eqnref{eq:negloglike_ll}.

\subsubsection{Regularization, learning and bias weights} \label{subsub:reg-learn-bias}
Optimization of \eqnref{eq:negloglike_cwf} is jointly non-convex in $\bm A$ and $\bm B$, but convex for $\bm A$ with $\bm B$ fixed, and vice versa. In practice that means we can only converge to a local minimum. Introducing regularization into the problem alleviates some non-convexity by excluding some local minima from the feasible set and additionally helps controlling overfitting. \cite{Menon2011} suggests an $\ell_2$ norm penalty, thereby effectively \textit{smoothing} the latent factors:
\begin{align}
\min_{\bm A, \bm B} \quad \Omega_{\ell_2}(\bm A,\bm B) - \sum_{(i,j) \in \mathcal{O}} C_{ij} \log(p^{MF}_{ij}) + (V_{ij} - C_{ij})\log(1 - p^{MF}_{ij}), \label{eq:negloglike_cwf_regl2}
\end{align}
where $\Omega_{\ell_2}(\bm A, \bm B) = \lambda(\sum_{i=1}^I ||\bm \alpha_i||_2^2 + \sum_{j=1}^J ||\bm \beta_j||_2^2)$. In this work we also try optimization with an $\ell_1$ norm regularizer:
\begin{align}
	\min_{\bm A, \bm B} \quad \Omega_{\ell_1}(\bm A,\bm B) - \sum_{(i,j) \in \mathcal{O}} C_{ij} \log(p^{MF}_{ij}) + (V_{ij} - C_{ij})\log(1 - p^{MF}_{ij}), \label{eq:negloglike_cwf_regl1}
\end{align}
with $\Omega_{\ell_1}(\bm A, \bm B)=\lambda_\alpha \sum_{i=1}^M |\bm \alpha_i|_1 +
\lambda_\beta \sum_{j=1}^N |\bm \beta_j|_1$, thereby promoting \textit{sparse} latent features.

For the $\ell_2$ regularized problem \eqnref{eq:negloglike_cwf_regl2}, a batch solver such as L-BFGS (see \cite[Chapter 9]{Nocedal1999}) can be invoked. For the $\ell_1$ regularized problem \eqnref{eq:negloglike_cwf_regl1}, special care must be taken due to non-differentiability. The quasi-newton method OWL-QN \cite{Andrew2007} can be used instead in this setting. For really large problems, an on-line learning framework, such as \textit{stochastic gradient descend} (SGD) is more scalable; again requiring special handling of the $\ell_1$ regularizer; see \cite{Tsuruoka2009} for details.

In general with classification problems with skewed class probabilities, \textit{e.g.}, observing many more non-clicks than clicks, we can add bias terms to capture such baseline effects. We follow the suggestion from \cite{Menon2010} and add separate bias terms for each row and column object, \textit{i.e.}, in our case per-page and per-banner bias weights. Hence, without loss of generality, when we refer to $\bm \alpha_i$ and $\bm \beta_j$, we assume they have been appended $[\alpha_0, 1]^T$ and $[1, \beta_0]^T$, respectively, thereby catering for the biases. Furthermore, when we speak of a rank $k$ latent feature model, we actually refer to a rank $k+2$ model consisting of $k$ latent features as well as the two bias dimensions.

\subsection{Feature-based response prediction}\label{sub:logreg}
A different approach to response prediction is a model based on explicit features available in each observation. In the case of click-through data, such information could for instance be attributes exposed by the browser (e.g., \textit{browser} name and version, \textit{OS}, \textit{Screen} resolution, etc.), \textit{time-of-day}, \textit{day-of-week} as well as \textit{user profiles} based on particular user's previous engagements with pages, banners, and with the ad server in general.

Again, we can use logistic regression to learn a model of binary labels: For $d = 1,...,D$ observations we introduce \textit{feature vectors}, $\bm x_d$, and model the probability of click given features with the logistic function, \textit{i.e.}, $p_d^{LR}=p^{LR}(click | \bm \omega_{LR})=\sigma(\bm \omega_{LR}^T \bm x_d)$. The optimization problem for learning the weights $\bm \omega_{LR}$ becomes
\begin{align}
	\min_{\bm \omega_{LR}} \quad \Omega_{\ell_1}(\bm \omega_{LR}) -  \sum_{d=1}^D y_d\log(p_d^{LR}) + (1-y_d)\log(1 - p_d^{LR}), \label{eq:negloglike_ll_reg}
\end{align}
where $\Omega_{\ell_1}(\bm \omega_{LR})=\lambda_{LR} |\bm \omega_{LR}|_1$ is added to control overfitting and produce sparse solutions. As discussed in Section \ref{subsub:reg-learn-bias}, adding bias terms can account for skewed target distributions, and may be included in this type of model, \textit{e.g.}, as a global intercept, by appending an all-one feature to all observations. Alternatively, if we want to mimic the per-page and per-banner biases of the latent factor model, we do so by including the page indices and banner indices encoded as one-of-$M$ and one-of-$N$ binary vectors, respectively, in the feature vectors. 

\subsection{Combining models}\label{sub:combined}
With dyadic response prediction as introduced in Section \ref{sub:cwfact}, the model can be extended to take into account \textit{side-information} available to each dyadic observation. \textit{I.e.}, introducing an order-3 tensor $\bm X$ with entries $\bm x_{ij} = \bm X_{ij:}$ being the feature vectors of side-information available to the $(i,j)$ dyad, we follow \cite{Menon2011} and model the confidence-weighted factorization with side-information as $p_{ij}^{SI}=p^{SI}(click|\bm\alpha_i,\bm\beta_j,\bm\omega_{SI}) = \sigma(\bm\alpha_i^T\bm\beta_j + \bm\omega_{SI}^T\bm x_{ij})$.

Learning such a model by jointly optimizing both $\bm \alpha$, $\bm \beta$, and $\bm \omega_{SI}$, is non-convex and may result in bad local minima \cite{Menon2010}. To avoid such bad minima, \cite{Menon2010, Menon2011} suggest a simple heuristic; first learn a latent-feature model as the one detailed in Section \ref{sub:cwfact}, then train the side-information model as in Section \ref{sub:logreg}, but given the log-odds ($\bm \alpha_i^T \bm \beta_j$) from the latent-feature model as input. \textit{I.e.}, $p_{ij}^{SI}$ can be rewritten as
\begin{align}
	p_{ij}^{SI}= \sigma([\bm 1;\bm\omega_{SI}]^T [\bm \alpha_i^T\bm \beta_j; \bm x_{ij}]),\label{eq:combined}
\end{align}
hence, having first learned $\bm\alpha_i^T\bm\beta_j$ c.f. Section \ref{sub:cwfact}, $\bm\omega_{SI}$ can be learned by extending the input features with the log-odds of the latent-feature model and fixing the corresponding weights to one. This training heuristic is a type of \textit{residual fitting}, where the side-information model is learning from the differences between observed data and the predictions by the latent-feature model.

In practice we have found the above procedure to be insufficient for obtaining the best performance. Instead we need to \textit{alternate} between fitting the latent features and fitting the side-information model, each while holding the predictions of the other model as fixed. This leaves how to train the latent feature model using the current side-information model prediction as fixed parameters open. Therefore we in the following show how this can be achieved.


For \eqnref{eq:negloglike_cwf_regl2}, which we will use as the working example, the observations are summarized for each unique dyad, $(i,j)$, in terms of the click and non-click counts, regardless of the side-information in each of those observations. Therefore we now address the question: Given $\bm \omega_{LR}$ from Section \ref{sub:logreg}, how do we obtain the quantities $\bm\omega_{SI}^T\bm x_{ij}$ in \eqnref{eq:combined}?

Initially, we define the notation $\bm x^{(i,j)}_d$ as indexing the $d^{th}$ explicit feature vector involving the dyad $(i,j)$. Hence, for dyad $(i,j)$ there are $V_{ij}$ (potentially) different feature vectors $\bm x^{(i,j)}_d$, $d=1,...V_{ij}$, involved. Assuming a model learned on the explicit features alone according to $p_d^{LR}$ from Section \ref{sub:logreg}, the overall predicted click-through rate for the observations involving dyad $(i,j)$ becomes
\begin{align}
	p_{ij}^{LR} := \frac{1}{V_{ij}} \sum_{d=1}^{V_{ij}} \sigma(\bm \omega_{LR}^T \bm x_d^{(i,j)}),\label{eq:lr_pred_ctr}
\end{align}
which is obvious from the fact that the sum calculates the predicted number of clicks and $V_{ij}$ is the empirical number of observations. \textit{I.e.}, \eqnref{eq:lr_pred_ctr} is just the average predicted click-through rate taken over the observations involving dyad $(i,j)$. Using this result we can now make sure the combined model yields $p_{ij}^{LR}$ when either $\bm \alpha_i = \bm 0$ or $\bm \beta_j = \bm 0$ (or both) by fixing the term $\bm \omega_{SI} \bm x_{ij}$ according to the log-odds of $p_{ij}^{LR}$. Hence,
\begin{align}
	\bm \omega_{SI}^T \bm x_{ij} = \log(p_{ij}^{LR}) - \log(1 - p_{ij}^{LR})
\end{align}
should be used as fixed inputs while learning the latent-factor model and thus accounts for the predictions of the feature-based model the same way as bias terms account for baseline effects.




\section{Data and experiments}

We will run experiments on datasets extracted from ad transaction logs in Adform, an international online advertising technology provider. Due to the sequential nature of these data, we will report results from training a model on 7 consecutive days and then testing on the 8$^{\text{th}}$. For measuring performance we report area under the ROC curve (AUC) scores as well as the logistic loss, evaluated on the held-out data (i.e., the last day). We evaluate different instantiations of the model over a period of in total 23 test days each using the previous 7 days for training and therefore can also report on the consistency of the results.

The data consists of observations labeled either click or not click and in each observation the domain and the banner id are always known. The additional features, that are features for the side-information, include various categorical variables all encoded as one-of-K. These include the web browsers \textit{UserAgent} string (a weak fingerprint of a user), an indicator vector of the top-50k (for a single country) websites the user has visited (\textit{URLs visited}) the past 30 days, the full URL of the site being visited (top-50k indicator vector, per country), a binned frequency of the times the user has clicked before (never, low, mid, high), as well as cross-features of the above mentioned and each banner id, thereby tailoring coefficients for each ad. The resulting number of features ($P$) is between 500k-600k and the number of positive observations is around 250k ($N_1$), i.e., the problem is overcomplete in features and thus $\ell_1$ on the side-information model is added as a means of feature selection. The negative class $N_0$ is down-sampled by a factor of 100 bringing it down to around 1.5M-2.5M. The resulting model can be corrected in the intercept \cite{King2001}. In our experience down-sampling the negative class this drastically and calibrating the model by intercept correction does not impact downstream performance.

We train different variations of the models to investigate in particular the usefulness of latent features in addition to a model using only explicit features. The different models are:
\begin{description}
	\item[LR] Logistic regression on the side-information alone. The corresponding regularization strength we call $\lambda_{LR}$.
	\item[LFL$_0$] The latent feature log-linear model using only the bias features, i.e., corresponding to a logistic regression for the two indicator features for domain and banner, respectively. The corresponding regularization weights we refer to as $\lambda_{\alpha_0}$ and $\lambda_{\beta_0}$, but as we describe later, in practice we use the same weight, $\lambda_0$, for both.
	\item[LFL$_K$] The latent feature log-linear model with $K>=0$ including bias features. The corresponding regularization weights we call $\lambda_{\alpha}$ and $\lambda_{\beta}$, which we also find in practice can be set to be equal, and for this introduce the weight $\lambda_{\alpha\beta}$.
	\item[LR+LFL$_K$] The combined model with $K \geq 0$ for the LFL model combined with the side-information model.
\end{description}

\subsection{Tuning hyper-parameters}
The combined model with both latent features and side-information we dub LR+LFL$_K$. This model has up to 5 (!) hyper-parameters that need tuning by cross-validation: $(\lambda_{LR}, \lambda_{\alpha_0}, \lambda_{\beta_0}, \lambda_{\alpha}, \lambda_{\beta})$. \cite{Menon2010} does not report whether they use individual $\lambda_{\alpha}$, $\lambda_{\beta}$, $\lambda_{\alpha_0}$, and $\lambda_{\beta_0}$ weights, but we consider this highly infeasible. What we have found to be most effective, is to use the same weight $\lambda_{\alpha\beta}$ for the latent dimensions as well as a shared bias weight $\lambda_0$, which narrows the search space down to three hyper-parameters that must be tuned.

Tuning three hyper-parameters is still a cumbersome task, in particular for large datasets, where an exhaustive search for a reasonable grid of three parameters becomes too time consuming. Instead we have had success using the following heuristic strategy for tuning of these parameters:
\begin{enumerate}
	\item First run experiments for the logistic regression model alone and find a suitable regularization weight $\lambda_{LR}$.
	\item Run experiments for the LFL$_0$ model (i.e., bias weights only) and find a suitable $\lambda_0$.
	\item Run experiments for a number of LFL$_K$ models with $K>0$, with bias weights regularized by $\lambda_0$ fixed from (2), and find a suitable $\lambda_{\alpha\beta}$.
	\item Finally, train the combined LFL+LR$_K$ model with different $K \geq 0$ and $\lambda_0$ fixed, but varying $\lambda_{\alpha\beta}$ as well as $\lambda_{LR}$ both in the \textit{neighborhood} of the values found in (1) and (3). If the results indicate performance could be improved in any direction away from that region, we run more experiments with the hyper-parameters set in that direction.
\end{enumerate}
To verify the validity of this approach, we have run experiments with the hyper-parameters set to their optimal settings as per the above procedure, and varying one at the time, including separate weights for the latent features and biases. In this way we do not find any increase in the performance along any single direction in the space of hyper-parameters.


\section{Results and discussion}

\subsection{Validation set results and initialization}
We use the first 8 days, i.e., train on 7, test on the 8$^{\text{th}}$, to find a reasonable range of hyper-parameters that we will test over the entire period. I.e., we use the first test day of a total of 23 days (30 days worth of data, where the first 7 are only used for training) as our \textit{validation set}. At the same we initialize models with different parameters, that we use for \textit{warm starting} the training on subsequent data. In the following we provide our results where we are testing performance on a single day, thereby gaining insights into both hyper-parameter values, model order ($K$) and regularization type ($\ell_1$ and $\ell_2$) for the latent features.

\begin{figure}
	\centering
	\begin{subfigure}[b]{0.49\textwidth}
		\includegraphics[width=1\textwidth]{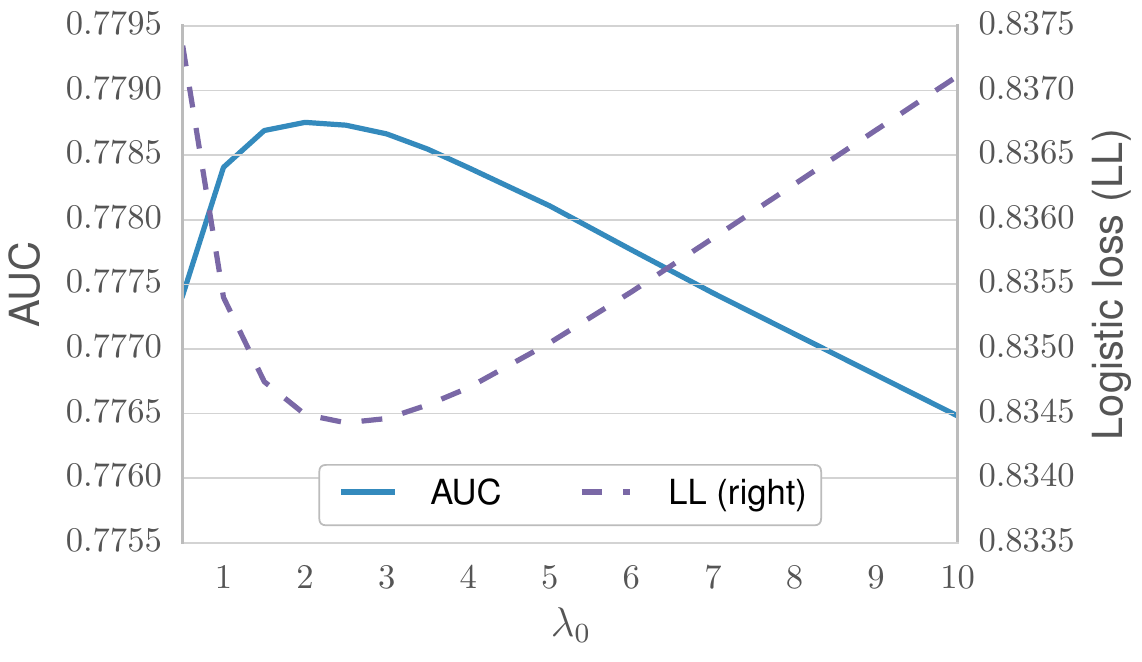}
		\caption{}
	\end{subfigure}\\
	
	\begin{subfigure}[b]{0.49\textwidth}
		\includegraphics[width=1\textwidth]{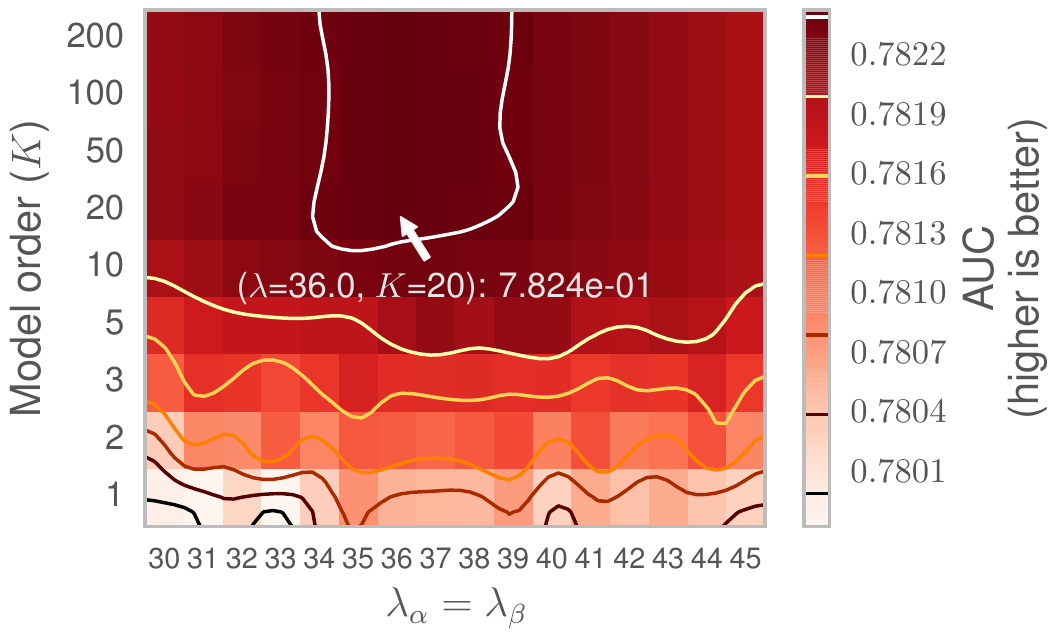}
		\caption{}
	\end{subfigure}
	\hfill
	\begin{subfigure}[b]{0.49\textwidth}
		\includegraphics[width=1\textwidth]{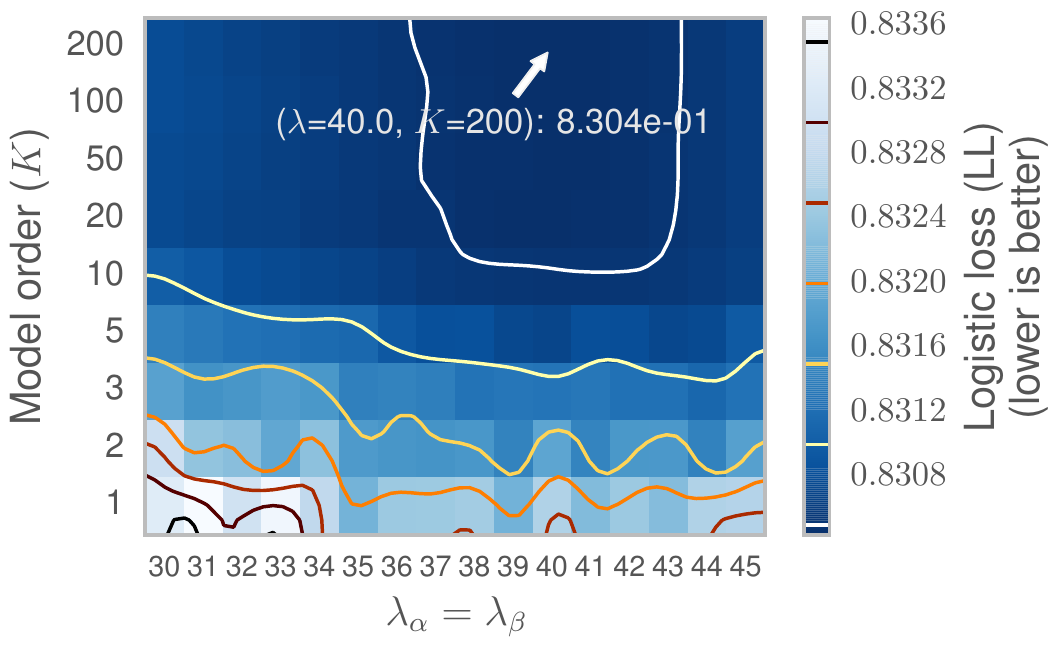}
		\caption{}
	\end{subfigure}
	\caption{Results using $\ell_2$ regularization for LFL$_K$ modeling on a single day with varying regularization strengths and model orders. (a) A sweep in over the LFL$_0$ regularization strength $\lambda_0$ in the vicinity of the optimum. (b) Using $\lambda_0=3.0$ the AUC is plotted as intensities in a grid with varying model orders and the shared regularization strength $\lambda_{\alpha \beta}$ for the latent dimensions. The annotations marks the optimum. (c) Same as (b), except for logistic loss, i.e., the lower the better.}
	\label{fig:l2_lflonly_cv}
\end{figure}

In \figref{fig:l2_lflonly_cv} we show the results using $\ell_2$ regularization (see \eqnref{eq:negloglike_cwf_regl2}) and varying $\lambda_0$ with an LFL$_0$ model (a) and $\lambda_{\alpha\beta}$ in (b-c) with $\lambda_0=3.0$ fixed, different model orders and no side-information model. Results are not shown for experiments where $\lambda_0$ and $\lambda_{\alpha\beta}$ were varied in larger grids, i.e., these plots focus of where the performances peak. What we also learn from these plots (b-c), is that higher model orders are advantageous, but that this increase levels off from between $K=5$ to $K=20$. This is in contrast to \cite{Menon2011} reporting $K\geq200$ being advantageous. We have also run experiments not shown here with $K=100$ and $K=200$ and seen no further increase, and if anything at all, a slight decrease in performance.

\begin{figure}
	\centering
	\begin{subfigure}[b]{0.49\textwidth}
		\includegraphics[width=1\textwidth]{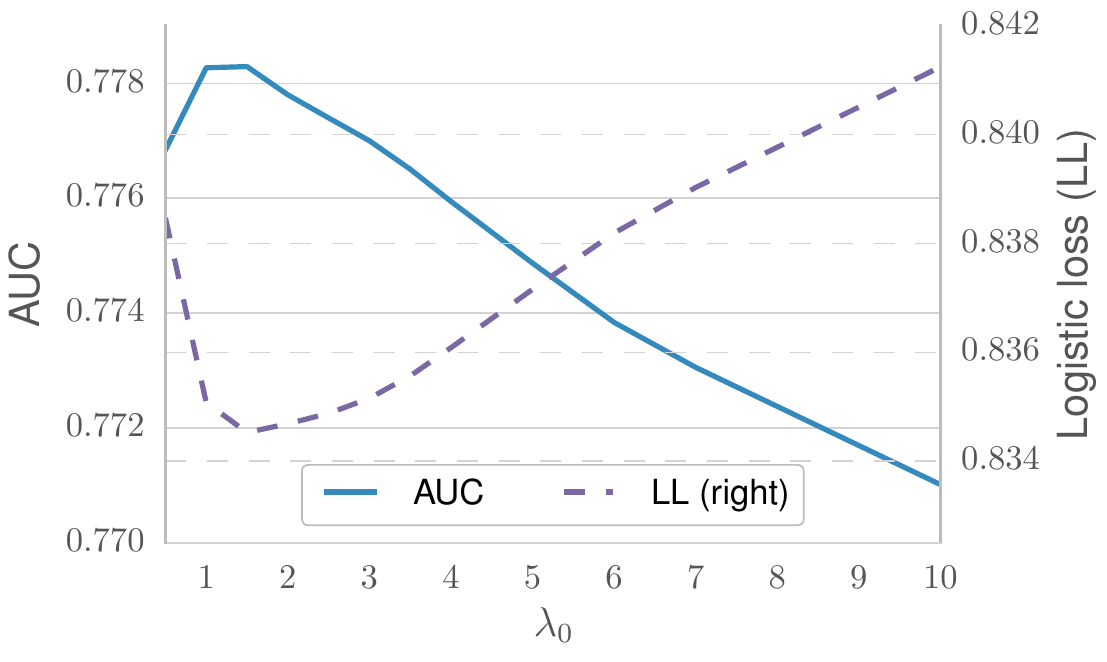}
		\caption{}
	\end{subfigure}\\
	
	\begin{subfigure}[b]{0.49\textwidth}
		\includegraphics[width=1\textwidth]{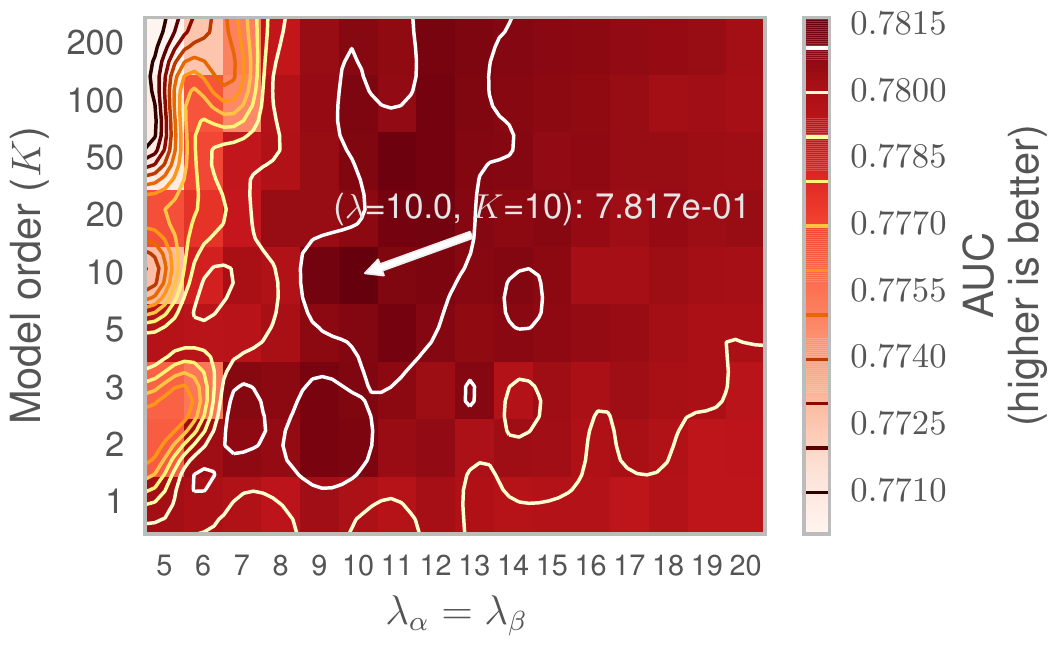}
		\caption{}
	\end{subfigure}
	\hfill
	\begin{subfigure}[b]{0.49\textwidth}
		\includegraphics[width=1\textwidth]{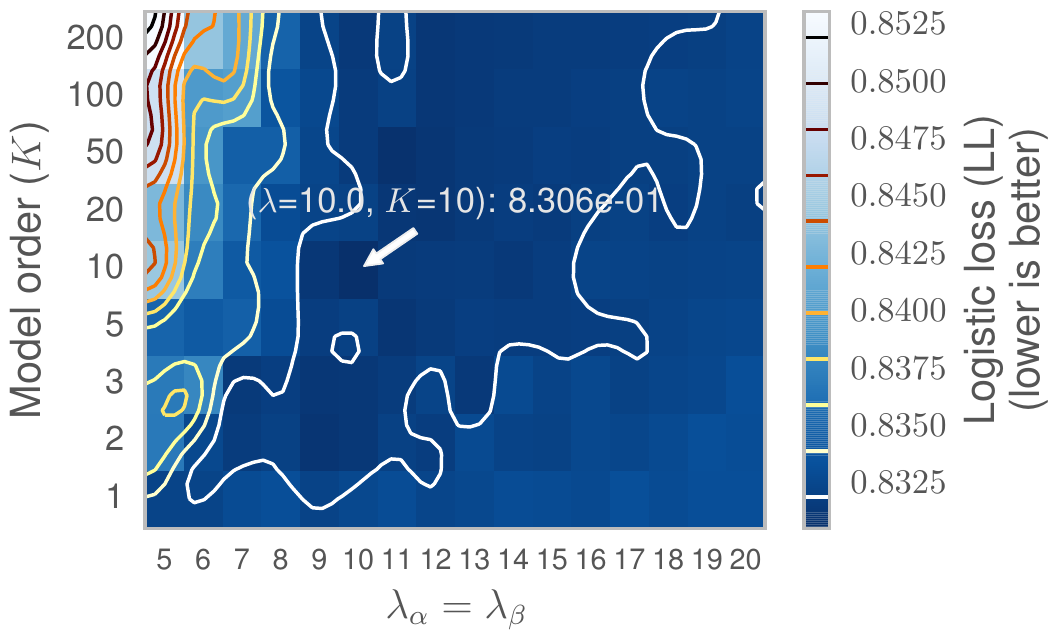}
		\caption{}
	\end{subfigure}
	\caption{Results using $\ell_1$ regularization for LFL$_K$ modeling on a single day with varying regularization strengths and model orders. Also see the caption for \figref{fig:l2_lflonly_cv}.}
	\label{fig:l1_lflonly_cv}
\end{figure}

The same experiments using $\ell_1$ regularization are summarized in \figref{fig:l1_lflonly_cv}. Both the experiments for the bias regularization $\lambda_0$ (a) as well as those for the latent factors (b-c) do not show as good performances as in the case of $\ell_2$ regularization and the advantage of adding latent dimensions is harder to distinguish. Furthermore the regions of interest seem more concentrated, i.e., the optima are more peaked. This leads us to the conclusion, that smoothness in the latent dimensions ($\ell_2$) is preferable to sparsity ($\ell_1$) and thus we do not report further results using $\ell_1$ regularization.

\begin{figure}
	\centering
	\begin{subfigure}[b]{1\textwidth}
		\includegraphics[width=1\textwidth]{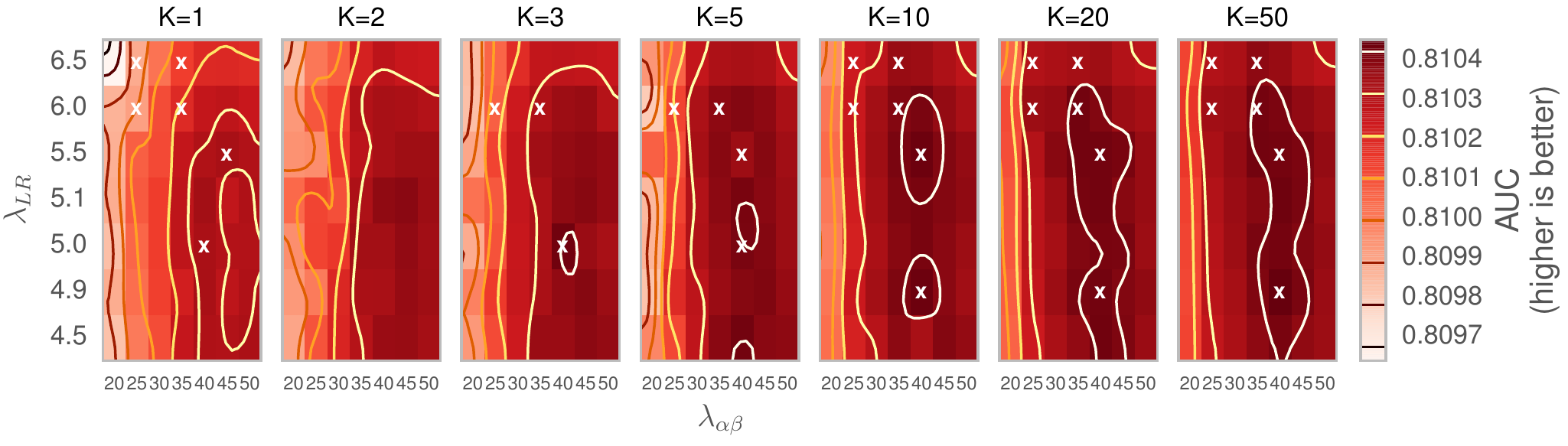}
	\end{subfigure}
	\begin{subfigure}[b]{1\textwidth}
		\includegraphics[width=1\textwidth]{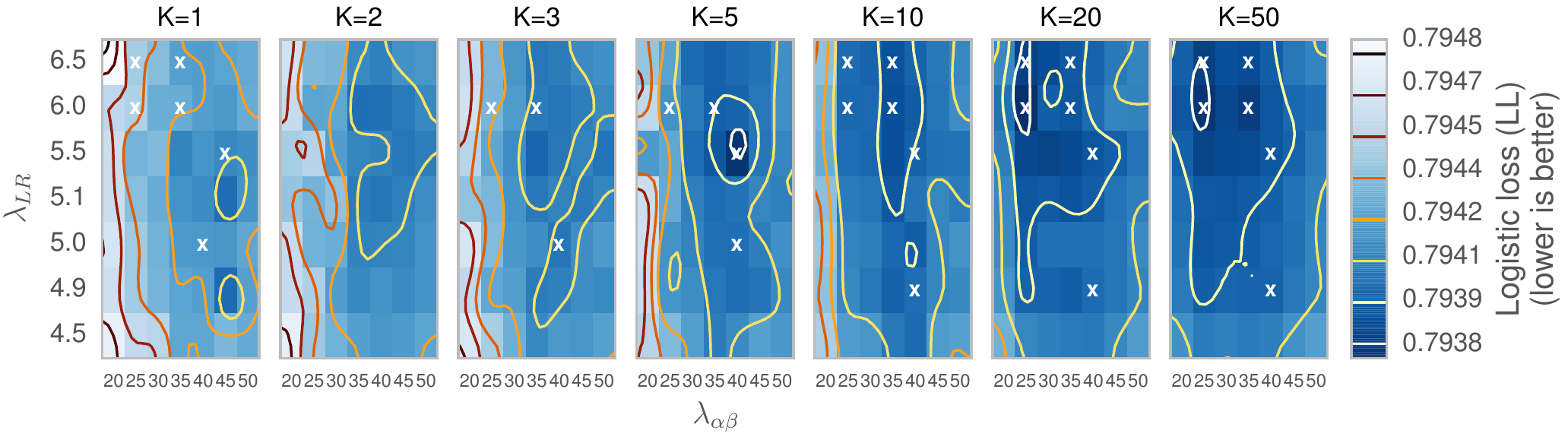}
	\end{subfigure}
	\caption{Results using the $\ell_2$ regularized version for LR+LFL$_{K}$ on the first day and varying regularization strengths as well as model orders. $\lambda_0=3.0$ remains fixed. In the top we show AUC intensities and in the bottom logistic losses. Little x's are used to mark those specific configurations that we run experiments with across all the test days.}
	\label{fig:lfllr_cv}
\end{figure}

In \figref{fig:lfllr_cv} we show experiments with varying $\lambda_{LR}$, $\lambda_{\alpha\beta}$ as well as different models orders $K$ for combined LR+LFL$_K$ models. We confirm a trend towards better performance using higher $K$, but again saturating beyond $K=5$. We further notice that peak performances in terms of AUC do not necessarily agree completely with those for LL. There may be other explanations as to why that is, but we believe this is a consequence of the LL being sensitive to probabilities being improperly calibrated, while the AUC is not. Inspection of the different models seem to confirm this; where the models perform better in terms of logistic loss, the predicted click-through rate for the test set is (slightly) closer to the empirical, than for those models which maximize AUC. We expect that a post-calibration of the models beyond just an intercept correction could be beneficial for the reported logistic losses, but also note that this would not change the AUCs.

\begin{figure}
	\centering
	\includegraphics[width=0.49\textwidth]{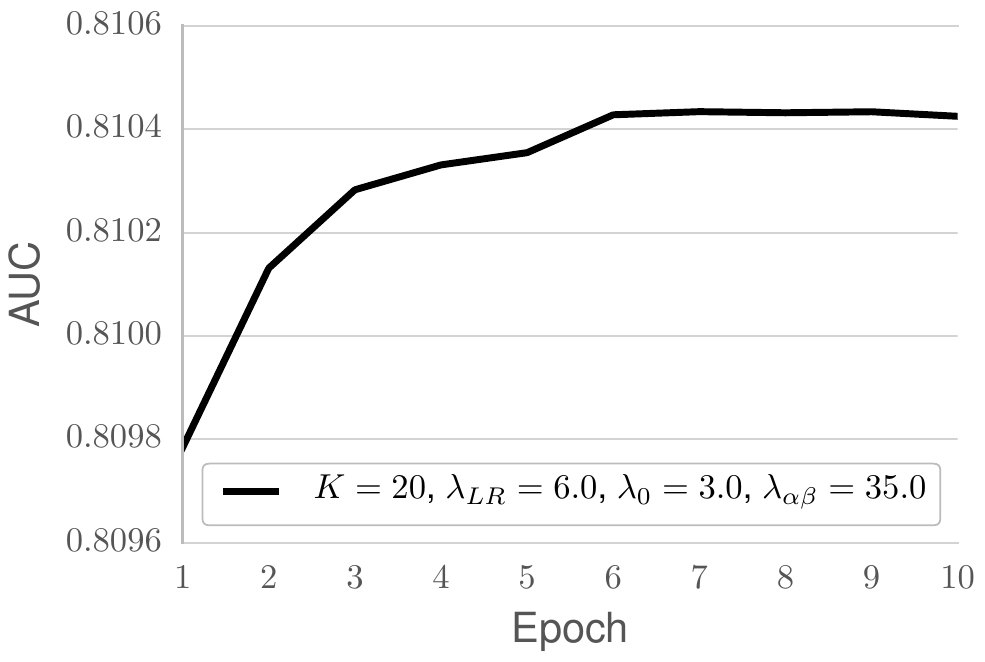}
	\caption{An example of a run of the LR+LFL$_K$ model with alternating updates to the latent and the side-information coefficients, which illustrates the generic level-off in performance (here AUC), as we run more epochs.}
	\label{fig:lfllr_alternate_auc}
\end{figure}

As mentioned in \secref{sub:combined}, we find that alternating between fitting the latent model and the side-information model is necessary. For the experiments \figref{fig:lfllr_cv}, we have alternated 7 times which we have confirmed in practice ensures the performance has leveled off. An example supporting this claim is shown in \figref{fig:lfllr_alternate_auc} and serves to illustrate the general observation we make in all our experiments.

\subsection{Results on 22 consecutive days}

With just a subset of LR+LFL$_K$ models hand-picked (marked by little x's in \figref{fig:lfllr_cv}) from the experiments on the first test day, we run experiments on a daily basis while initializing with the models from the previous day. This sequential learning process reflects how modeling would also be run in production at Adform and by warm starting the models in the previous days' coefficients, we do not expect that running multiple epochs of alternated fitting is required, i.e., this only needs to be done once for initialization.

In the following, the AUCs and logistic losses we report are daily averages of the performances for each banner. As opposed to the performances over an entire test data set that we have reported up until now, making daily averages per banner prevents the performance numbers from being entirely dominated by a single or a few banners, and instead assigns per-banner performances equal weights.

\begin{figure}
	\centering
	\begin{subfigure}[b]{1\textwidth}
		\includegraphics[width=1\textwidth]{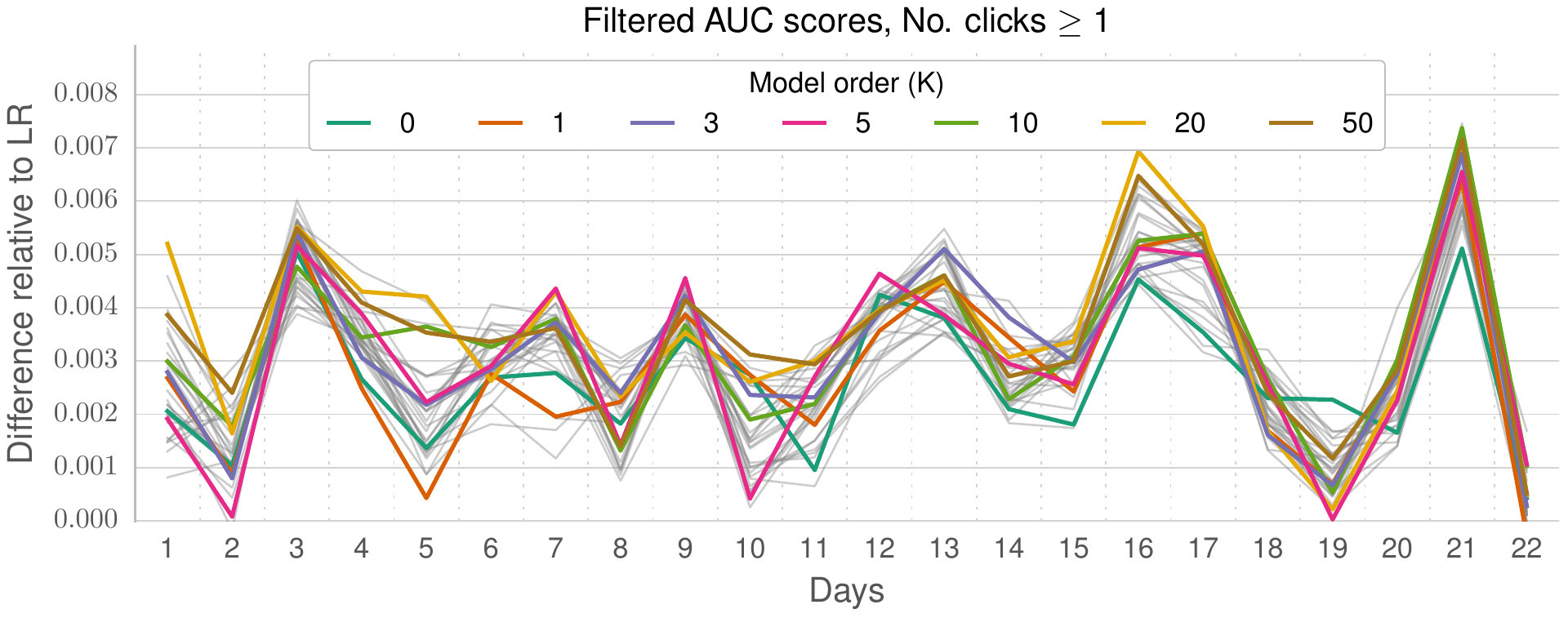}
	\end{subfigure}\\
	\bigskip
	\begin{subfigure}[b]{1\textwidth}
		\includegraphics[width=1\textwidth]{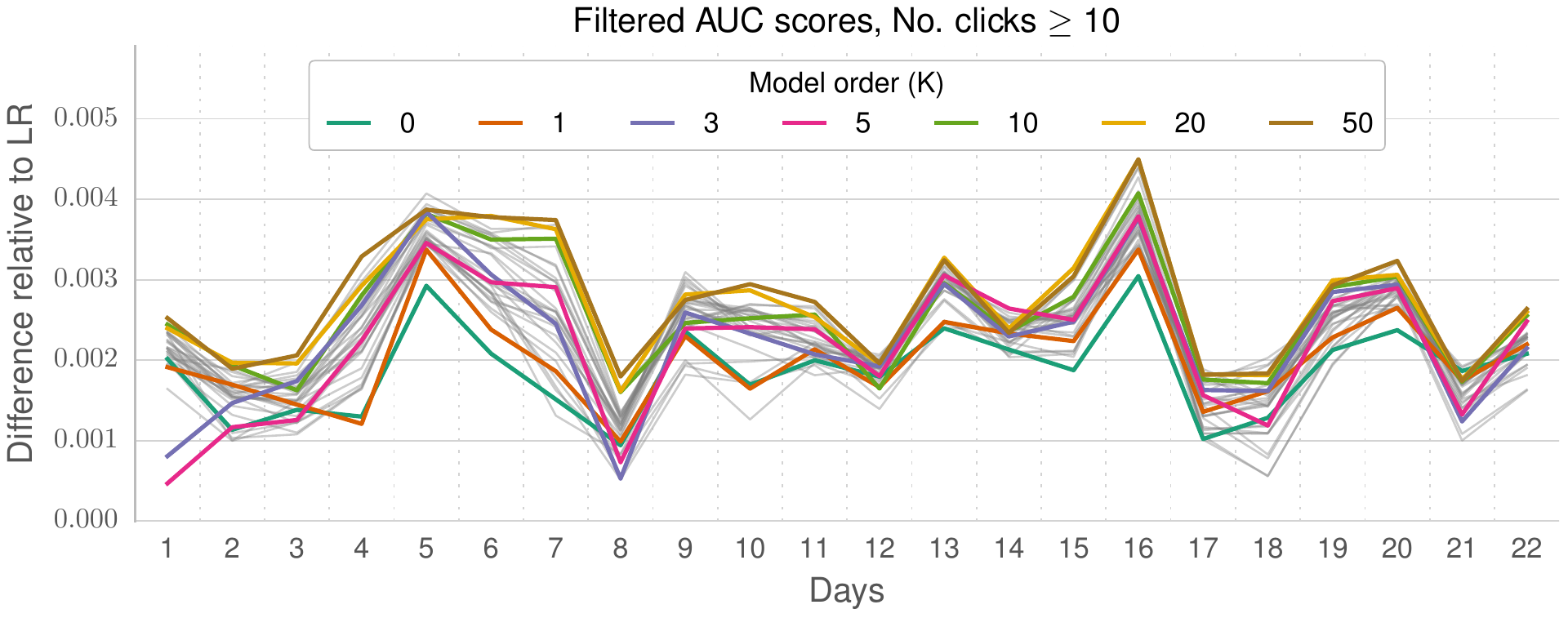}
	\end{subfigure}
	\caption{Daily average AUC differences (i.e., increase) for LR+LFL$_K$ models using the optimal settings for different model-orders (colored lines) relative to the side-information model alone. Shaded, gray lines in the background trace all of the different configurations tested. In the above the averages include every banner with 1 or more clicks on a day (between 900-1000 banners qualify each day), while in the lower all the banners on a particular day with less than 10 clicks is filtered from the averages (between 400-500 banners qualify each day), hence decreasing the variance in the measures.}
	\label{fig:lfllr_alldays_auc_time}
\end{figure}

Reporting performances based on slices of data per banner further allows analysis of under which circumstances the latent feature models add a statistical significant improvements. In \figref{fig:lfllr_alldays_auc_time} we show the difference in AUC banner averages per day in the total of 22 days we use for testing. The upper shows the performances for all the banners with 1 or more clicks in each test set (day), while the lower is averaged daily performances for only the banners with 10 or more clicks. It is apparent from these two figures, that AUC scores based on very few clicks add significant variance to the daily averages and the difference between the model orders is hard to spot. We also note, that since we cannot evaluate AUCs score for banners without clicks in the test set, these are ignored entirely. For logistic loss, however, we can still report a performance for banners without clicks in the test. The LR model used as the reference (0.0) in \figref{fig:lfllr_alldays_auc_time} uses $\lambda_{LR}=4.0$, which we found as optimal over the entire 22 day period testing a grid from $2.0$ to $7.0$ in increments of $0.5$.

\begin{figure}
	\centering
	\begin{subfigure}[b]{1\textwidth}
		\includegraphics[width=1\textwidth]{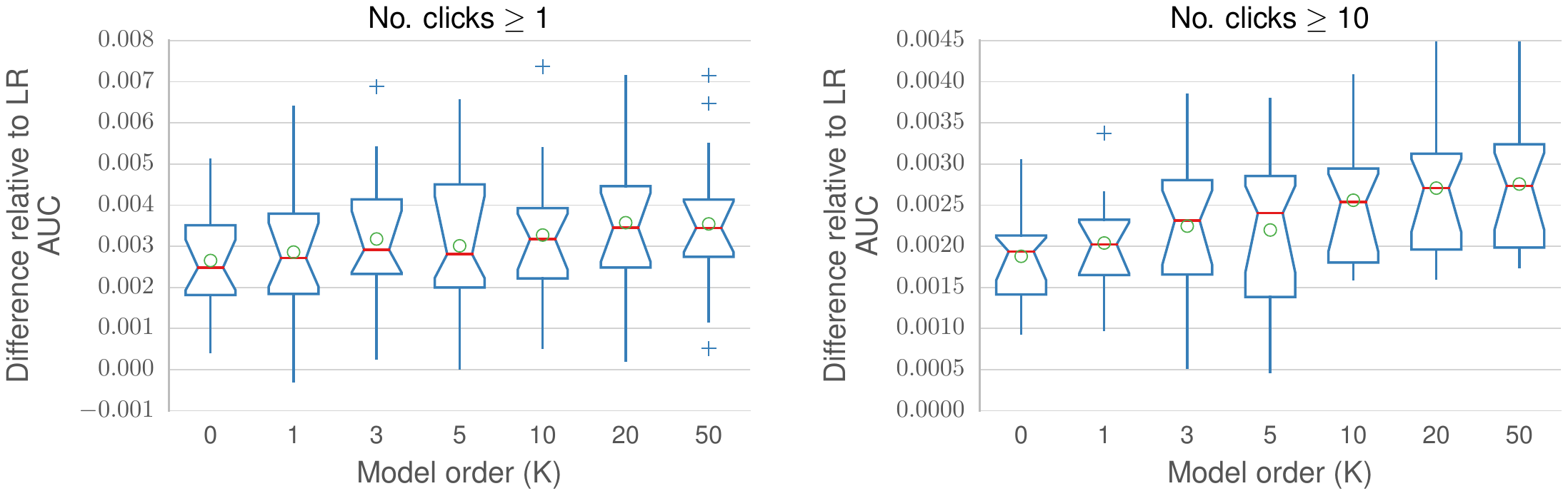}
	\end{subfigure}\\
	\bigskip
	\begin{subfigure}[b]{1\textwidth}
		\includegraphics[width=1\textwidth]{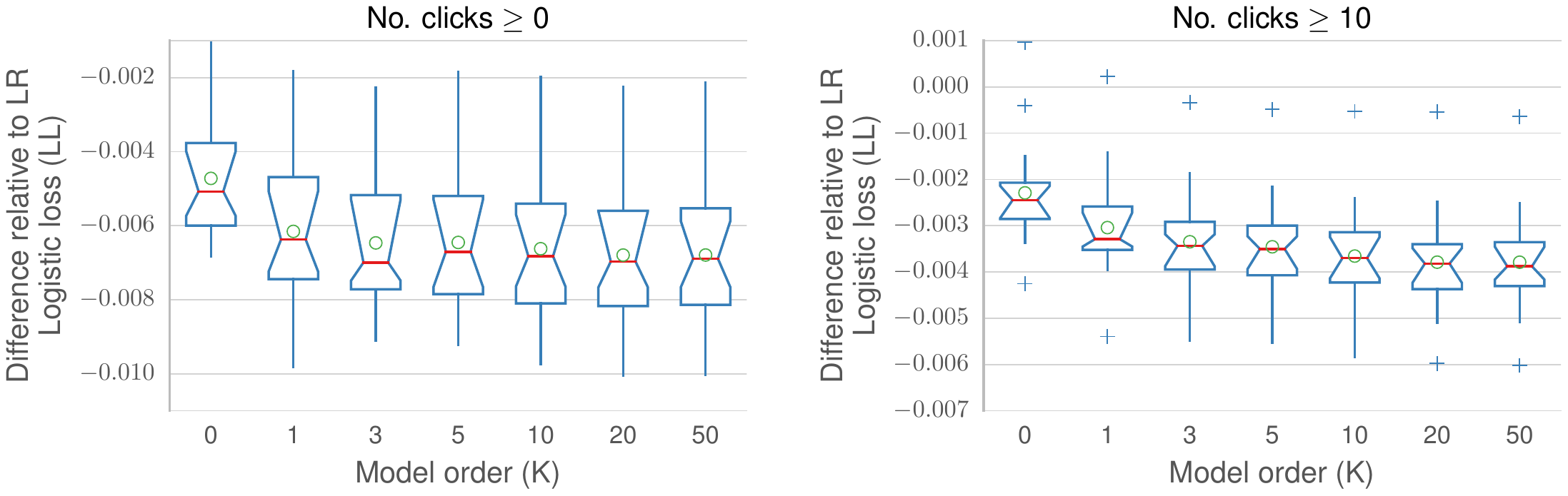}
	\end{subfigure}
	\caption{\textbf{Top row}: Box plots of the relative AUC differences corresponding to each of the models also displayed in \figref{fig:lfllr_alldays_auc_time}. \textbf{Bottom row}: Box plots also for the relative logistic loss differences with the only difference being that in the left-most box plot, the losses for \textit{all} banners each day are included in the averaging (1100-1200 daily). \textbf{Legend}: Boxes show $1^{\text{st}}$ and $3^{\text{rd}}$ quartiles, whiskers are 1.5 IQR (interquartile range), medians are red lines and green circles mark the means. Outliers are marked with pluses. The notches are 5000 bootstrap sample estimates of the 5\%-95\% confidence interval for the medians.}
	\label{fig:lfllr_alldays_box}
\end{figure}

In order to further quantify and investigate the impact different model orders has on performance, we summarize in \figref{fig:lfllr_alldays_box} the relative differences over the 22 test days in box plots. Again, we show performances relative to the side-information model and for different inclusion criteria, based on number of clicks in the test sets. In all cases, we see an increase in performance, as the model order is increased, and this increase levels off from $K=10$ to $K=50$. The notches on boxes are 5k sample bootstraps of the medians, hence based on these we can say something about the statistical significance of these results. I.e., non-overlapping notches correspond to $p<0.05$ for a two-tailed null-hypothesis test. First of all, all model orders, including $K=0$, improve performances compared to the side-information model alone.

For both the AUCs and the logistic losses we see wide confidence intervals on the medians, when banners with very few clicks ($<10$) per day are included. We still observe an increase in performance as the model order increases, but only in the case of logistic loss do the model orders $K=20$ and $K=50$ barely clear overlapping with the notches of $K=0$.

In the case of including only banners with more than 10 clicks in the summary statistics, the confidence intervals of the medians shrink, in particularly in the case of logistic loss. However, the relative gains (means and medians) are also slightly lower. I.e., there is a \textit{trend}, albeit barely statistically significant, that there are higher gains among the banners with few clicks in the test sets, than for those with more. Apart from this, there is now also statistically significant differences between the medians for the higher model orders and $K=0$; in the case of AUC this includes $K=20$ and $K=50$, and in the case of logistic loss, $K\geq 3$ are statistically better. 

It is worth noting that, regardless of the slice based on number of clicks in the test sets, the results agree that using the LR+LFL$_K$ model yields higher performance than the LR model alone.

For the results in \figref{fig:lfllr_alldays_box}, while we find evidence that supports that latent features improves click-through prediction, the question remains \textit{how much} this improves real-world performances. Indeed the increments which the latent features introduce in the two measures we report here seem very small. When measuring AUC scores, in particular, we are however not the first to report small, but significant improvements on the third decimal. As \citet{McMahan2011} (on web search ads) puts it:
\begin{quote}
\textit{The improvement is more significant than it first appears. A simple model with only features based on where the ads were shown achieves an AUC of nearly 0.80, and the inherent uncertainty in the clicks means that even predicting perfect probabilities would produce an AUC significantly less than 1.0, perhaps 0.85.}\cite[p.532]{McMahan2011}
\end{quote}
Our data as well as our experiences in web banner ads support this statement, and we often also identify new features or model changes with these low levels of improvement, but which however remain consistent.

Another possibility as an alternative to \textit{off-line} measures on held-out data, such AUC and logistic loss, is \textit{live} A/B testing. Yet, before taking new features, models or technology into production, a prerequisite to us at least, is to demonstrate consistent off-line data performance improvements. For the present work, we have not had the opportunity to test it live.

\section{Conclusion}
In this work we have reviewed a method for click-through rate prediction which combines collaborative filtering and matrix factorization with a side-information model and fuses the outputs to proper probabilities in $[0,1]$. We have provided details about this particular setup that are not found elsewhere and shared results from numerous experiments highlighting both the strengths and the weaknesses of the approach.

We test the model on multiple consecutive days of click-through data from Adform ad transaction logs in a manner which reflects a real-world pipeline and show that predictive performance can be increased using higher-order latent dimensions. We do see a level-off in the performances for $\approx K\geq 20$, whereas $K\geq 200$ was suggested in another work \cite{Menon2011}, but may be due to differences in the data sets; in particular how many side-information features are available and used.

Our numerous experiments detail a very involved phase for finding proper regions for the various hyper-parameters of the combined model. This is particularly complicated, since the latent feature model and the side-information model need to be trained in several alternating steps, for each combination of hyper-parameters. This we think is one of the most severe weaknesses of this modeling approach. We circumvent some of the complexity of finding good hyper-parameters by using shared regularization strengths for both entities of the latent model and demonstrate, that in a sequential learning pipeline, it is only for initialization of the model, i.e., on the first training set, that we need multiple alternating steps.

For future studies, it would be particularly useful if the hyper-parameters could instead be inferred from data. Yet, as we also show in our results, the objective differences (i.e., the \textit{evidence}) that separate good models from the bad, are small, hence we expect any technique, such as Type II maximum likelihood, would be struggling to properly navigate such a landscape.

\bibliographystyle{plainnourl}
\bibliography{ref}

\end{document}